# FEATURE EXTRACTION METHODS FOR COLOR IMAGE SIMILARITY

R.Venkata Ramana Chary[1], Dr.D.Rajya Lakshmi[2] and Dr. K.V.N Sunitha[3]

[1]Padmasri Dr.B.V Raju Institute of Technology, Hyderabad, India
rvrchary@gmail.com
[2]GITAM Institute of Technology, Visakhapatnam, India
rdavuluri@yahoo.com
[3]G. Narayanamma Institute of Technology and Science Hyderabad, India
k.v.n.sunitha@gmail.com

## ABSTRACT

*In this paper  we concentrated on image retrieval system in early days many  user interactive systems performed with  basic concepts but such systems are not reaching to the user specifications and not attracted to the user so a lot of research interest in recent years with new specifications , recent approaches have user is interested in  friendly interacted  methods are expecting ,  many are concentrated for improvement in all methods. In this proposed system we focus on the retrieval of images within a large image collection based on color projections and different mathematical approaches are introduced and applied  for retrieval of images. before Appling proposed methods  images are sub grouping using threshold values,  in this paper  R G B color  combinations considered for retrieval of images ,in  proposed methods are implemented and results are included ,through results it is observed that  we  obtaining  efficient results comparatively previous and existing.*

## 1. INTRODUCTION

The term content based image retrieval (CBIR)[1] is the application of computer vision techniques to the image related problems. Experiments into automatic retrieval of images from a large collection database are based on the colors, shapes and texture and image features [2]. At current stage, there is a gap between low-level features of the retrieval system and the high-level semantic[6] concepts of the user, called semantic gap. Compared with text-IR, this problem result in very poor performance of CBIR system .All techniques and algorithms that are used originate from fields such as statistics, pattern recognition[4][5] and computer vision.

Content-Based Image Retrieval (CBIR) is according to the user-supplied in the bottom characteristics, directly find out images containing specific content from the image library The basic process: First of all, do appropriate pre-processing of images like size and image transformation and noise reduction is taking place, and then extract image characteristics needed from the image according to the contents of images to keep in the database. When we retrieve to identify the image , extract the corresponding features[6][7] from a known image  and then retrieve the image database to identify the images which are similar with it, also we can give some of the characteristics based on a query  requirement, then retrieve out the required images based on the given suitable  values. In the whole retrieval process, feature extraction is essential; it is closely related to all aspects of the feature, such as color, shape, texture and space.

## 1.1. Color Image

A color image is a combination of some basic colors. In MATLAB breaks each individual pixel of a color image (termed 'true color') down into Red, Green and Blue values. We are going to get as a result, for the entire image is 3 matrices, each one representing color features. The three matrices are arranging in sequential order, next to each other creating a 3 dimensional m by n by 3 matrixes. An image which has a height of 5 pixels and width of 10 pixels the resulting in MATLAB would be a 5 by 10 by 3 matrixes for a true color image.

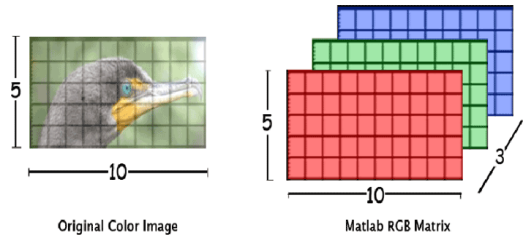

Figure 1. Color image and RGB matrix.

## 1.2 Color Panel

The following picture showing color panels projecting the color components in image is representing in following pictures 2 and 3.

R=RGB(:,:,1);  G=RGB(:,:,2); B=RGB(:,:,3);

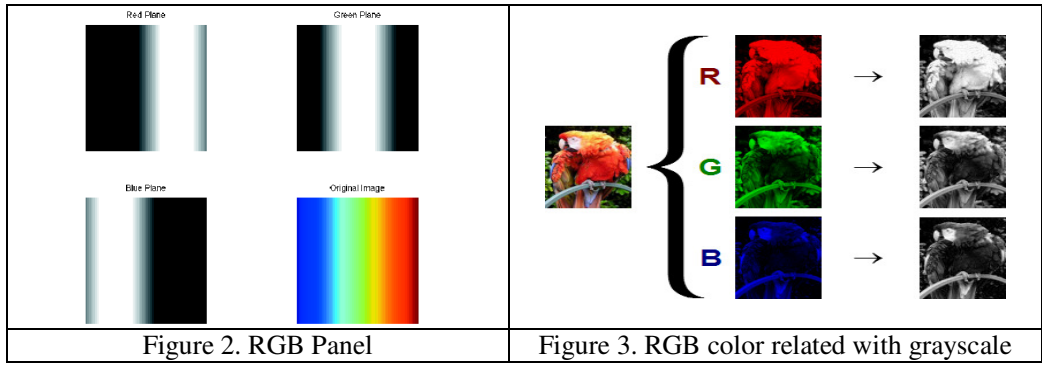

| Figure 2. RGB Panel | Figure 3. RGB color related with grayscale |
|---|---|

## 2. PROBLEM DESCRIPTION

In image retrieval system for searching, browsing, and retrieving images from a large database of images [14]. Most conventional and common methods of image retrieval utilize some method of adding metadata such as tokens, captioning , keywords, or descriptions to the images so that retrieval can be performed over the annotation words. Some systems are working with lower level features, Manual image annotation is time-consuming, laborious and expensive. To address this, many researchers are proposed on automatic[13] user friendly image retrievals using different methods.

Content-based means that the search will analyze the actual contents of the image. The term 'content' in this context might refer to colors, shapes, textures, or any other information that can be derived from the image itself. Without the ability to examine image content, searches must rely on metadata such as captions or keywords, which may be laborious or expensive to produce. In this paper proposed methods are providing the best solution in large image set.

# 3. PROPOSED SYSTEM

➤  Proposed system uses combinations of color feature to overcome the problem description.

➤  Proposed system is implemented and concentrated on visual contents of an image such as color, shape, texture and spatial layouts.

➤  Proposed system selected 10000 image databases with common feature values.

➤  Proposed system extracted all images features separately R, G, B values for problem solving.

➤ Proposed system implemented features like color histogram, color projections

➤ Mathematical approaches like mean, median and standard deviation are proposed for efficient retrieval

➤  Proposed work provides platform to extract images from the database using user query method.

# 4. WORK IMPLEMENTATION

## 4.1 Image retrieval is implemented in the following steps

Image retrieval is implemented in two different phases. One is new image insertion with features in database and other one is searching a new image in available database.

**Step 1:-** All ten thousand images are taken in to working directory of MATLAB.

**Step 2**:- Using MATLAB programming all image features in R, G, B color projection values are extracted and stored in database using specified programming methods.

**Step 3**:-Threshold calculation is taken for categorizing the images into a similar feature groups. In this step, threshold value is computed based on the histogram calculation.

If the image is a color image it will convert into gray color when calculated by the sum of all bins  in image histogram. Figure 4 is one example for threshold values of images.

following steps computes the  specifying threshold for image

ifisrgb(Image)              GImage=rgb2gray(Image);

p12=imhist(GImage);     threshold=sum(sum(p12))

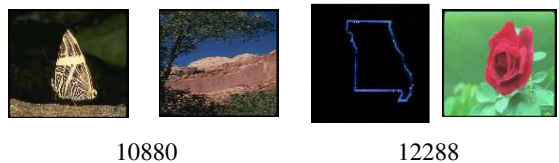

10880                    12288

Figure 4.

Using these methods, all images are categorizing into N number of Groups

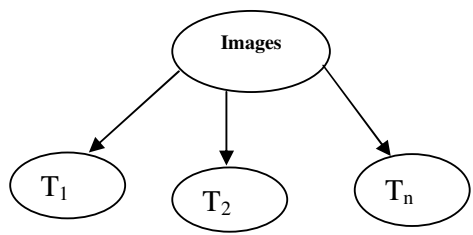

**Step 4**:-In usual methods, image color values are storing in matrix form. Using Image matrix, all R, G, B components in image are extracted and separated into three different array forms. ( Red , Green , Blue )

Red(M ,J)      =Image(M,J,1);
Green(M,J)     =Image(M,J,2);
Blue(M ,J)     =Image(M,J,3);

**Step 5** :- Using the feature vectors, each image color wise means are computed . In this method row and overall image mean are computed and stored into the database. Based on all this features different computing methods are formulated.

mean_r=mean2(Red);
mean_g=mean2(Green);
mean_b=mean2(Blue);

```
median_r=median((median(Red))');
median_g=median((median(Green))');
median_b=median((median(Blue))');
std_r=std((std(Red,0,1))',0,1);
std_g=std((std(Green,0,1))',0,1);
std_b=std((std(Blue,0,1))',0,1);
```

**Step 6**:- Query image is selecting based on user choice and verifying threshold value if the threshold values are equal and then proposed methods are implementing

**Step 7**:- Using the image feature vectors various retrieval methods are proposed. In each method, two different working group sets are identified. One is large image set group and second one is small set of image group.

All Images = { $T_1$, $T_2$, $T_3$ ...... $T_n$ }

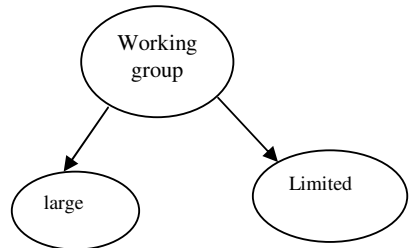

Large =$T_{equalent\ set}${ T is threshold value and all images belongs to same T group }
Limit =$T_{equalent\ set\ with\ less\ images}$

## 4.2 Proposed Method (PM)

In each method, two image feature values are verified from available database files, after that comparing equalities of query image (QI) and target image (TI). TI may be similar or equal or not equal. So based on the difference factor(DF), images are identified.

DF= Constant {this value is proposed by user}

**PM 1**:- Selecting only one color mean value from image vector database means are compared in the range of DF.

**PM 2**:- Selecting two colors mean values from the image feature database if the means are equal or similar values within the range of DF. Two colors mean are like RG, RB, and GB.

**PM 3**:- Selecting three colors mean values from the image feature database if the means are equal or similar values within the range of DF.

**PM 4**:- Selecting three color median values from the image feature database if the medians are equal or similar values within the range of DF

**PM 5**:- Selecting the different color standard deviation methods (SD) are applied for image comparisons. This method has given a good result comparatively. Other proposed method results are discussed in section 5.

## 5. RESULTS

### 5.1. Experimental Results

Query is applied by using mean values

| p.name | R_mean | Picture | p.name | R_mean | Picture |
|---|---|---|---|---|---|
| 998.jpg | 63.36859 | 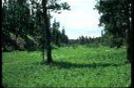 | 992.jpg | 83.37225 | 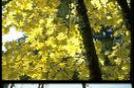 |
| 997.jpg | 65.08289 | 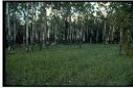 | 991.jpg | 84.91315 | 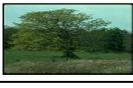 |
| 995.jpg | 75.65936 | 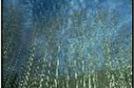 | 996.jpg | 86.83167 | 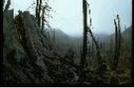 |
| 993.jpg | 82.29895 | 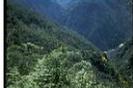 | 994.jpg | 110.7877 | 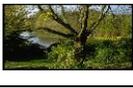 |
| Figure 5 : R_Mean Pictures | | | | | |

In above figure 5  it is showing that if the image is searching based on red_mean type top 8 pictures are displaying in table order .in above example DF value is 25 from pic 993.jpg to R_mean value is 75.65 remaining pictures -25 and +25 mean values considered .

| Table 1: Mean methods on 8 pictures and retrieval ranking | | | | | | | |
|---|---|---|---|---|---|---|---|
| PIC | R MEAN | G | B | RG AVG | RB | GB | RGB |
| 991.jpg | 6 | 6 | 5 | 6 | 4 | 6 | 4 |
| 992.jpg | 5 | 5 | 7 | 5 | 8 | 5 | 8 |
| 993.jpg | 4 | 7 | 8 | 7 | 6 | 7 | 7 |
| 994.jpg | 8 | 8 | 2 | 1 | 5 | 8 | 6 |
| 995.jpg | 3 | 3 | 1 | 3 | 2 | 3 | 2 |
| 996.jpg | 7 | 4 | 6 | 4 | 7 | 4 | 5 |
| 997.jpg | 2 | 1 | 3 | 1 | 1 | 1 | 1 |
| 998.jpg | 1 | 2 | 4 | 2 | 3 | 2 | 3 |

.

In the  Table 1 is showing that picture names and color means of R,G,B  ,two color combination averages and final RGB is calculated values represented  8 images .in above  table all proposed methods  are  applied based on that ranks are showing  .

In the above tables based on the color mean combination methods results are showing that almost 75 to 90 % of picture is looking similar .At the same time we can notify that  the results for  single color to combination of colors  features if  combining similar  images are coming together .this we can observe  through ranking(1 to 8) , in this  observation we can conclude that instead of using single color projection we can select different combinations of color methods applying we can achieve good performance.

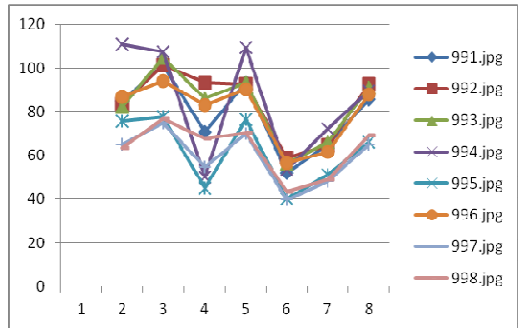

Figure 6.

In above graph(**Figure 6**) images relations in various proposed methods are indicated .(2 is r_mean,3 is g_mean,4 is b_mean,5 is rg_avg, 6is rb_avg,7is gb_avg,8 is rgb_avg means.)

**5.2** Second proposed method based on median .

Median red applied for selected set of images in that following order is performed

| Query image | | **7374** | |
|---|---|---|---|
| 7375 | | 7379 | |
| 7377 | | 7378 | |
| 7376 | | 7372 | |
| 7374 | | 7370 | |
| 7378 | | 7379 | |
| 7373 | | 7371 | |

Figure 7: Median red projection applied

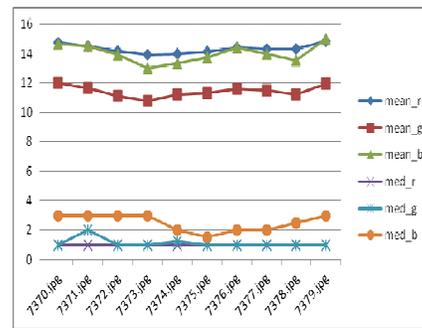

Figure 8 mean median on image set

In above graph(Figure 8) all pictures mean and median values are projected .

**5.3 Selection of Huge database using median method:** median red projection method is applied for the large image set, following pictures order is displayed .

| 7373 | | 5978 | | 5910 | | 5803 | |
|---|---|---|---|---|---|---|---|
| 7374 | | 5977 | | 7370 | | 5804 | |
| 7375 | | 5976 | | 7379 | | 9051 | |

| 7372 | | 5926 | | 5926 | | |
|------|--|------|--|------|--|--|
| 5938 | | 5805 | | 5822 | | |
| Figure 9:Median red applied on Large set data | | | | | | |

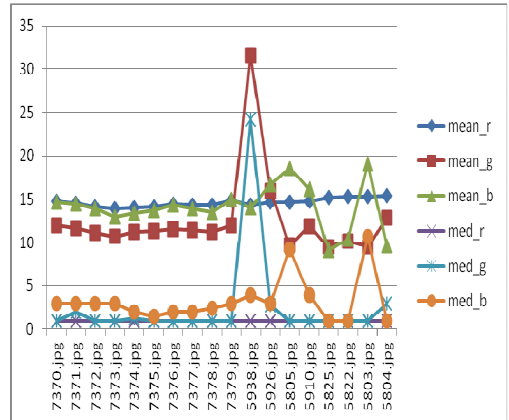

Figure 10: mean and median on image set

Large database is considered for retrieval of images .through graph(Figure 10) mean and medians method variations we can observe .at the same time pictures similarity[7][11] also changing so using proposal methods useful to find pictures similarities with efficient way.

### 5.4 Standard deviation green is applied in selected set of images in which following order is performed

| 3037 | 3039 | 3038 | 3036 |
|------|------|------|------|
| 3041 | 3044 | 3046 | 3048 |
| 3049 | 3054 | 3073 | 3075 |
| 3076 | 3084 | | |

Figure 11: SD method  green projection

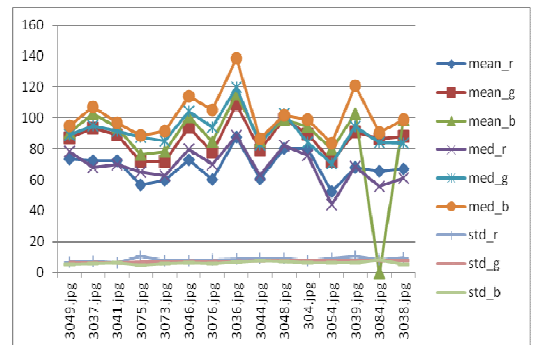

Figure 12:mean med and SD on image set green

In Figure 11 Standard Deviation method is applied sequence of images founded .in figure 12 graph showing that all method and relations between the methods on specified set of the images.

**5.5 Standard deviation red is applied in selected set of images in that  following order is performed**

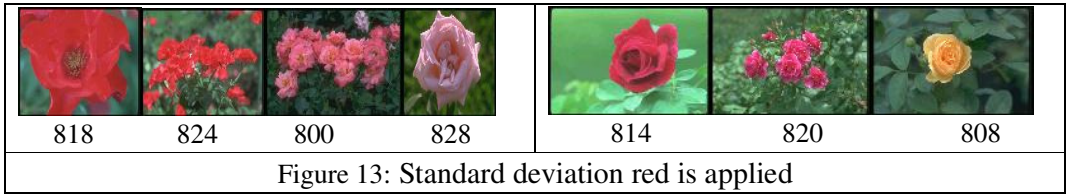

| | | |
|---|---|---|
| 818    824    800    828 | 814    820    808 |

Figure 13: Standard deviation red is applied

Table 2: Data values for comparing SD red  values

| pic name | Mea nr | Mea ng | Me anb | Me d_r | Me d_g | Me d_b | St d_r | St d_g | St d_b |
|---|---|---|---|---|---|---|---|---|---|
| **818.jpg** | 124 | 68 | 58 | 138 | 56 | 58 | 10.31 | 8.92 | 7.43 |
| **824.jpg** | 102 | 78 | 62 | 86 | 78 | 65 | 11.00 | 6.39 | 6.72 |
| **800.jpg** | 96 | 70 | 64 | 81 | 69 | 54 | 11.20 | 5.74 | 7.13 |
| **828.jpg** | 99 | 93 | 78 | 83 | 91 | 75 | 11.42 | 6.10 | 7.17 |
| **814.jpg** | 114 | 125 | 91 | 109 | 155 | 95 | 13.44 | 10.58 | 7.11 |
| **820.jpg** | 88 | 93 | 68 | 80 | 97 | 69 | 14.47 | 6.43 | 6.80 |
| **808.jpg** | 69 | 79 | 61 | 63 | 80 | 64 | 22.85 | 9.36 | 6.38 |

Figure 14:mean ,median and SD blue method on    set of images

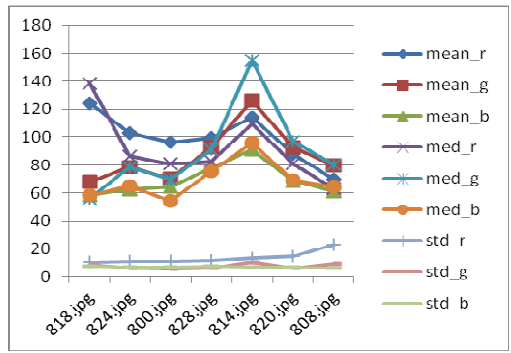

Table 3: Data values for  SD red  ranking values

| Fnapi cname | Mea nr | Mea ng | Me anb | Me d_r | Me d_g | Me d_b | St d_r | St d_g | St d_b |
|---|---|---|---|---|---|---|---|---|---|
| **818.jpg** | 7 | 1 | 1 | 7 | 1 | 2 | 1 | 5 | 7 |
| **824.jpg** | 5 | 3 | 3 | 5 | 3 | 4 | 2 | 3 | 2 |
| **800.jpg** | 3 | 2 | 4 | 3 | 2 | 1 | 3 | 1 | 6 |

| 828.jpg | 4 | 5 | 6 | 4 | 5 | 6 | 4 | 2 | 5 |
|---------|---|---|---|---|---|---|---|---|---|
| 814.jpg | 6 | 7 | 7 | 6 | 7 | 7 | 5 | 7 | 4 |
| 820.jpg | 2 | 6 | 5 | 2 | 6 | 5 | 6 | 4 | 3 |
| 808.jpg | 1 | 4 | 2 | 1 | 4 | 3 | 7 | 6 | 1 |

Above two tables are indicating feature values of  R G B colors of images .in the second table if 818.jpg image is searching according to red SD, ranks are specified from 1 to 7. Similarly if any other image is searching in database with different feature selection, images display in above rank order.

**5.6 Standard deviation blue** is applied in selected set of images in which following order is performed and all method relations are specified through a graph.

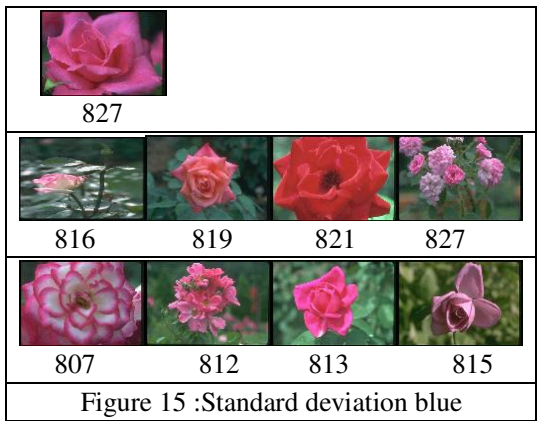

827

816      819      821      827

807      812      813      815

Figure 15 :Standard deviation blue

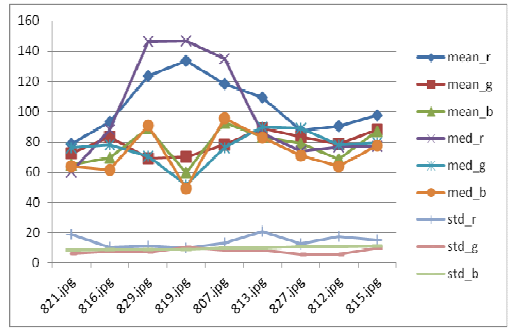

Figure 16:mean med and SD on blue

# 6. FUTURE ENHANCEMENT

Most Internet-based Content-Based Image Retrieval (CBIR) systems focus on different stock photo collections and do not address challenges of large specialized image collections and topics such as semantic[6] information retrieval by image content. in our research work we concentrated more than 10000 images with different categories of images .this work is helping towards large set of image retrieval applications. In proposed methods depending on the data set we can choose suitable methods and implement a good semantic[2] approaches in large set of images.

# 7. CONCLUSION

This paper proposes evaluation of image retrieval methods  All proposed  method are  shown different  comparison of the obtained results with other approach demonstrates that the proposed approach improves  accuracy rate comparatively existed  methods which is used to retrieve the images from the large database .in this work we compared total 10000 images with different categories. All suggested methods are helpful to perform the good results and based on query images what are the images retrieved from the database ,all  images showing vary similar and from method to method results are showing variation so best one is to select the combinations of colors mean with median and standard deviation , So we can expect  using proposed methods best performs and good results. Current techniques are based on low level features and there is a huge semantic gap . In future days more research work is required in semantically competent systems.

**Authors**

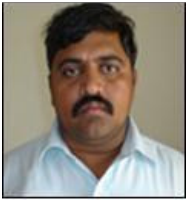

R.Vekata Ramana Chary is working as a Associate Professor in the Department of CSE at Padmasri Dr. B.V.Raju Institute of Technology, Narsapur, Medak, AP, India. Presently he is pursuing Ph.D from GITAM University, AP, India. His research interest is in Image Processing, Performance Measurements, Algorithms Analysis and Programming Techniques. He is having 15 years of teaching and research experience

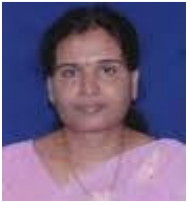

Dr. D. Rajya Lakshmi is working as Professor and HOD in the Department of IT at GITAM University, Visakhapatnam, AP, India. Her research areas include Image processing and Data mining. She is having about 17 years of teaching and research experience. Her PhD was awarded from JNTUH, Hyderabad, AP in the area of Image Processing

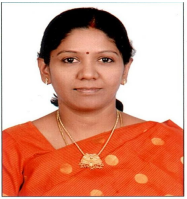

Dr. K.V.N. Sunitha is working as a Professor and HOD in Department of CSE at G. Narayanamma Institute Of Technology and Science (for WOMEN) in Hyderabad, AP, India. Her PhD was awarded from JNTUH Hyderabad AP. She is having 18 years of teaching and research experience in the Computer Science field. Her research interest is in Compilers, NLP, Speech Recognition, Computer Networks and Image Processing.